%% file: main.tex
\documentclass[runningheads]{llncs}

\usepackage{eccv}

\usepackage{eccvabbrv}

\usepackage{graphicx}
\usepackage{booktabs}

\usepackage[accsupp]{axessibility}  

\newcommand{\method}{3DPE\xspace}

\makeatletter
\newcommand{\thankssymbol}[1]{\textsuperscript{\@fnsymbol{#1}}}
\makeatother

\usepackage{hyperref}

\usepackage{orcidlink}

\begin{document}

\title{Real-time 3D-aware Portrait Editing \texorpdfstring{\\}~from a Single Image}

\titlerunning{3DPE}


\author{
    Qingyan Bai\inst{1, 2} \and
    Zifan Shi\inst{1} \and
    Yinghao Xu\inst{3} \and
    Hao Ouyang\inst{1, 2} \and \\
    Qiuyu Wang\inst{2} \and
    Ceyuan Yang\inst{4} \and
    Xuan Wang\inst{2} \and \\
    Gordon Wetzstein\inst{3} \and
    Yujun Shen\thanks{Corresponding authors.}\inst{2} \and
    Qifeng Chen\thankssymbol{1}\inst{1}
}

\authorrunning{Q.~Bai et al.}

\institute{
    The Hong Kong University of Science and Technology
    \and
    Ant Group \and Stanford University \and Shanghai AI Laboratory
}

\maketitle

\input{sections/0.abs}
\input{sections/1.introduction}
\input{sections/2.related}
\input{sections/3.method}
\input{sections/4.experiments}
\input{sections/5.conclusion}

\section*{Acknowledgements}
This project was supported by the National Key R\&D Program of China under grant number 2022ZD0161501.

%
%
\bibliographystyle{splncs04}
\bibliography{main}
\end{document}

%% file: sections/0.abs.tex
\begin{abstract}

This work presents \textbf{\method}, a practical method that can efficiently edit a face image following given prompts, like reference images or text descriptions, in a 3D-aware manner.
To this end, a lightweight module is distilled from a 3D portrait generator and a text-to-image model, which provide prior knowledge of face geometry and superior editing capability, respectively.
Such a design brings two compelling advantages over existing approaches.
First, our method achieves \textit{real-time} editing with a feedforward network (\textit{i.e.}, $\sim$0.04s per image), over 100$\times$ faster than the second competitor.
Second, thanks to the powerful priors, our module could focus on the learning of editing-related variations, such that it manages to handle various types of editing simultaneously in the training phase and further supports \textit{fast adaptation} to user-specified customized types of editing during inference (\textit{e.g.}, with $\sim$5min fine-tuning per style).
Project page can be found \href{https://ezioby.github.io/3dpe/}{here}.
\keywords{3D-aware portrait \and Efficient editing}

\end{abstract}

%% file: sections/1.introduction.tex
\section{Introduction}
\label{sec:intro}

Inferring the geometry and appearance from a single-view portrait image has become mature and practical~\cite{lin20223d, ko20233d,xie2023high, ide3d, narrate,live3d,gao2020portrait}, largely attributed to the utilization of priors in various 2D/3D generative models.
However, only performing geometry reconstruction is insufficient. 
The significance of 3D portrait editing, driven by user intentions, and the need for streamlined efficiency in the editing process has been steadily increasing.
This is particularly crucial in real-world applications such as AR/VR, 3D telepresence, and video conferencing, where real-time editing is often essential. 
Consequently, a key question arises: How can we effectively address the challenge of attaining high-fidelity portrait editing while ensuring real-time efficiency?

\begin{figure}[t]
  \centering
  \includegraphics[width=1.0\linewidth]{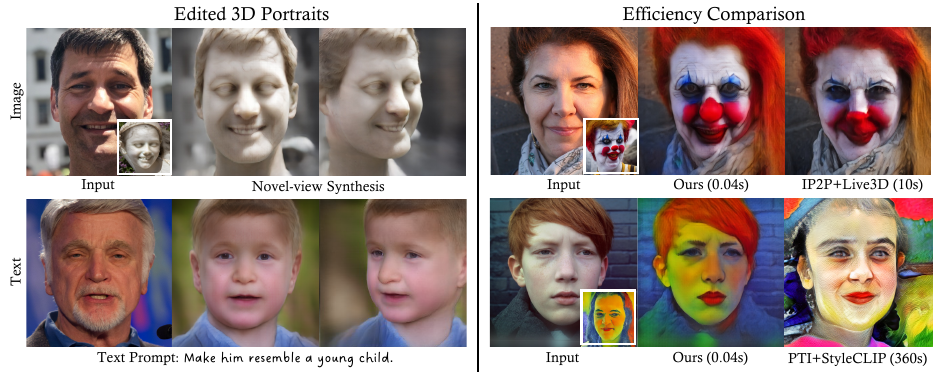}
  \caption{
    \textbf{Photorealistic editing results} produced by our proposed \textbf{\method}, which allows users to perform 3D-aware portrait editing using image or text prompts.
    In comparison with baseline methods, such as InstructPix2Pix (IP2P)~\cite{instructpix2pix}+Live3D~\cite{live3d} and PTI~\cite{pti}+StyleCLIP~\cite{styleclip} (details are illustrated in Sec.~\ref{sec:exp}), our approach accurately follows the guidance from reference prompts and maintains sufficiently better efficiency.
  }
  \label{fig:teaser}
\end{figure}

Traditional methods for 3D portrait editing~\cite{deng2020disentangled} typically rely on template facial models~\cite{blanz2023morphable,paysan20093d}, which has limitations in handling the substantial geometry changes because it often overlooks precise features like hair and beard.
Recent 3D GANs~\cite{graf,pigan,eg3d,3dsurvey,xia2023survey} show remarkable capabilities in generating high-fidelity 3D portraits.
They can serve as powerful generative priors for 3D portrait editing when coupled with GAN inversion techniques~\cite{zhu2016generative,idinvert,interfacegan,image2stylegan, image2stylegan++}.
However, these methods may encounter issues related to geometry distortion or exhibit slow speed, and the editing is constrained by the limited latent attributes.
Besides, 2D diffusion models can offer a strong prior with Score Distillation Sampling (SDS) loss~\cite{dreamfusion} for editing purposes~\cite{hertz2023delta}. 
Nonetheless, they often require step-by-step optimization and become a major bottleneck for real-time applications.

To this end, we present \textbf{\method}, a real-time \textbf{3D}-aware \textbf{P}ortrait \textbf{E}diting method driven by user-defined prompts.
As shown in~\cref{fig:teaser}, when provided with a single-view portrait image, our approach empowers a versatile range of editing styles with one model through flexible instructions, including images and texts.
We leverage the powerful 3D prior from a 3D-aware face generator and achieve a high-fidelity 3D reconstruction of the portrait image. 
Subsequently, we distill editing knowledge from a text-guided image editing model into a lightweight module integrated with the 3D-aware generator.
This module, characterized by its minimal computational cost, allows our method to maintain real-time inference and excel in handling various types of editing, while ensuring good 3D consistency.
An additional advantage is that our model supports customization through user-specified prompts with fast adaptation speed. 
This empowers users to build their own editing model at a minimal cost, enabling our method to cater to a broader audience.

Our method achieves real-time 3D-aware portrait editing through the utilization of a feedforward network, with a processing time of 40ms on a standard consumer GPU.
Additionally, we present a comprehensive evaluation of our method using various prompts both quantitatively and qualitatively. 
Our design choices are also validated through comparisons with ablated variants of our method.
Compared with baseline methods, our approach demonstrates superior 3D consistency, precise texture alignment, and a substantial improvement in inference time, as reflected by the evaluation metrics.
We demonstrate the versatility of our method by showcasing its capacity to perform a wide variety of edits, including text and image prompts, on portrait images.

In summary, the contributions of our work include:

\begin{itemize}
    \item We propose a lightweight module to distill knowledge from 3D GANs and diffusion models for 3D-aware editing from a single image. 
    Due to the minimal cost of the new module, our model maintains real-time performance.
    
    \item Our model supports fast adaptation to user-specified editing, requiring only 10 image pairs and 5 minutes for the adaptation.
    
    \item Our framework can accommodate various control signals, including text and image prompts.
\end{itemize}

%% file: sections/2.related.tex
\section{Related Work}
\label{sec:related_works}

\subsection{Generative Face Priors}
Generative models aim at modeling the underlying distribution of the data, containing a wealth of prior knowledge.
Recently, 3D GANs~\cite{graf,pigan,eg3d,volumegan,epigraf,voxgraf,geod,pof3d,shadegan, stylenerf, stylesdf, wang2023benchmarking, tran2024voodoo} are mostly adopted to learn 3D faces from single-view image dataset. 
The interior rich domain-specific geometry priors enable various face-related applications such as image editing~\cite{live3d,pix2nerf,fenerf,ide3d, zhou2021cips, yang2022pastiche} and domain adaptation~\cite{jin2022dr, kim2023datid, zhang2023deformtoon3d, abdal20233davatargan, nitzan2023domain}.
Large-scale diffusion models~\cite{stablediffusion}, in contrast, encode knowledge of huge datasets and thus can provide general prior information.
Such priors are broadly leveraged for tasks such as image editing~\cite{sdedit, controlnet, plugandplay, instructpix2pix, cao2023masactrl}, customization~\cite{dreambooth, lora, liu2023cones, kumari2023multi}, and video editing~\cite{ouyang2023codef, liu2023video, ceylan2023pix2video, qi2023fatezero, wang2023zero, yang2023rerender, geyer2023tokenflow, chai2023stablevideo}.
In our approach, we capitalize on the advantages of both the geometry prior derived from 3D GANs and the broader editing prior offered by large-scale text-to-image diffusion models, instead of relying solely on a single type of prior.

\subsection{Portrait Editing from a Single Image }

Although many methods work well for reconstructing~\cite{guo2022perspective,gao2020portrait} or generating~\cite{eg3d} faces, editing has become a necessary interface to connect these methods with real-world applications.
Previously, portrait editing given a single-view image was mostly completed in 2D image space facilitated with GAN inversion~\cite{interfacegan, idinvert, image2stylegan, image2stylegan++,pti,sefa,ghfeat,bai2022high, harkonen2020ganspace}, which enables fast editing by exploring the trajectories in the GAN's latent space. 
Recent diffusion-based portrait editing~\cite{instructpix2pix, controlnet} can support various editing types with texts as guidance.
However, most of them are done in the 2D space, and thus there is no guarantee for the underlying 3D consistency.
Therefore, 3D-aware portrait editing is crucial to achieve the goal.
Some methods~\cite{fenerf,lin20223d,ide3d,narrate,xie2023high,yin20233d,hairnerf,li2023preim3d,jiang2023nerffacelighting} rely on the latent space of 3D GANs to perform editing through walking in the latent space, but are limited by the number of editable latent attributes.
A few methods, such as Instruct-NeRF2NeRF~\cite{instructnerf2nerf}, ClipFace~\cite{clipface}, and LENeRF~\cite{hyung2023local}, attempt to leverage the large-scale models~\cite{clip, stablediffusion} to edit on 3D representations with texts as guidance but require heavy iterative refinement of the edited results.
Similarly, InstructPix2NeRF~\cite{instructpix2nerf} leverages text-to-image models and text prompts to edit within the latent space of the GAN-based 3D generator. However, it is held back by the multi-step diffusion inference and the heavy structure of a GAN-based generator.
Ours, in contrast, benefits from the lightweight design of the module and achieves real-time editing with ease. 
With such a design, our model can also adapt fast to the user-defined editing prompts.
Moreover, texts cannot always illustrate the desired effects precisely, and sometimes an image prompt serves as better guidance for image editing~\cite{gal2022textualinv, dreambooth}.
Our method can support not only text prompts but also image prompts for editing, providing a more flexible and user-friendly interface.

%% file: sections/3.method.tex
\section{Method}
\label{sec:method}

Our approach takes a portrait image $\mathbf{I}$ and its camera pose \textbf{c}, obtained via the face pose estimator~\cite{deng2019accurate}. 
Additionally, the referenced prompts $\mathbf{P}$ (images or texts) serve as the editing instructions. 
The outcome of our model is an edited version of the 3D portraits characterized by NeRF in accordance with the prompts.
The edited image is denoted as $\mathbf{I}_{p}$.
Within our framework, we achieve this objective by leveraging the knowledge of 3D GANs and a text-guided image editing model, which is distilled into a lightweight module.
Following distillation, our method enables real-time 3D portrait editing and efficient adaptation to user-specified prompts.
In Sec.~\ref{method:prior}, we offer an illustration of 3D GANs and diffusion models.
The process of distilling these priors into a lightweight module is outlined in Sec.~\ref{method:distill}.
Finally, the details for model training and inference, along with the fast adaptation for customized prompts, are presented in Sec.~\ref{method:training}.

\begin{figure}[t]
    \centering
    \includegraphics[width=1.0\linewidth]{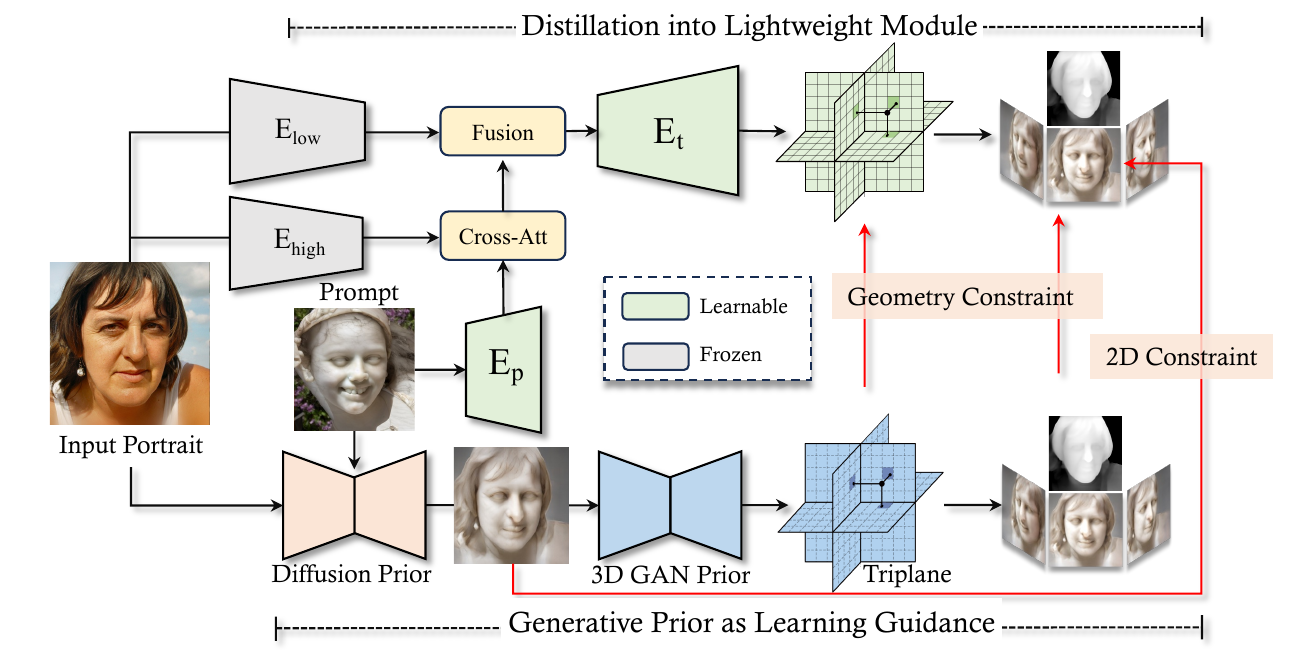}
    \caption{\textbf{Overview of our method}. 
    We distill the prior in the diffusion model and 3D GAN for real-time 3D-aware editing. Our approach is fine-tuned from Live3D~\cite{live3d}, where we extract features from the input portrait $\mathrm{I}$ using $\mathrm{E}_{high}(\cdot)$ and $\mathrm{E}_{low}(\cdot)$. The prompt embedding is generated with $\mathrm{E}_{p}(\cdot)$ and injected with the input features from $\mathrm{E}_{high}(\cdot)$ through a cross-attention mechanism. Our model is trained to mimic the output from the diffusion prior to acquire editing knowledge and enforce geometry constraints through triplane, multi-view images, and depth supervision from the 3D prior. In this context, InstructPix2Pix~\cite{instructpix2pix} and Live3D serve as the diffusion and 3D prior, respectively. It's noteworthy that only $\mathrm{E}_{p}(\cdot)$ and $\mathrm{E}_{t}(\cdot)$ are learnable during training, while all other parameters remain frozen.}
    \label{fig:frameowrk}
\end{figure}

\subsection{Preliminary}~\label{method:prior}

\noindent{\textbf{3D GAN Prior for Portrait Reconstruction}}.
The 3D-aware GANs showcase the ability to synthesize photorealistic 3D images using a collection of single-view images.
Notably, EG3D~\cite{eg3d} introduces an efficient triplane 3D representation, demonstrating high-quality 3D-aware image rendering.
Once trained, the generator of EG3D can be applied for single-image 3D reconstruction via GAN inversion.
However, existing 3D GAN inversion methods often encounter geometry distortion or exhibit slow inference speed.
To tackle these challenges, we utilize Live3D~\cite{live3d}, a state-of-the-art single-image portrait 3D reconstruction model built upon EG3D, preserving geometry quality while ensuring real-time performance.
Live3D employs a two-branch encoder, $\mathrm{E}_{high}(\cdot)$ and $\mathrm{E}_{low}(\cdot)$, to extract different resolution features from the input. 
It then utilizes a ViT-based $\mathrm{E}_{t}(\cdot)$ decoder~\cite{vit} to transform the fused encoder features into the triplane $\mathbf{T}$:
\begin{align}
    \mathbf{T} = \mathrm{E}_{t}(\mathrm{E}_{high}(\mathbf{I}), \mathrm{E}_{low}(\mathbf{I})).
\end{align}
This triplane $\mathbf{T}$ is subsequently used in conjunction with the volume rendering module and upsampler of EG3D to generate photorealistic view synthesis given the camera pose $\mathrm{c}$. For clarity, we use $\mathrm{R}(\cdot)$ to denote this process.

\noindent{\textbf{Diffusion Prior for Portrait Editing}}.
Only performing 3D reconstruction is insufficient as our goal is to edit portrait images. 
Large-scale diffusion models~\cite{stablediffusion}, trained on vast text-image pairs, can synthesize realistic photos with text input, offering powerful editing prior. 
Despite their effectiveness for face images, the editing process is often slow due to step-by-step optimization or multiple inferences in diffusion models. 
The scarcity of high-quality and multi-view consistent paired data significantly hinders the practical application of 3D-aware editing from a single image, which is hard to acquire with diffusion models.
Therefore, our goal is to distill editing knowledge from the diffusion model, integrate it with the 3D prior in Live3D into a lightweight module, and employ it for real-time portrait editing.

\begin{figure}[t]
  \centering
  \includegraphics[width=0.90\linewidth]{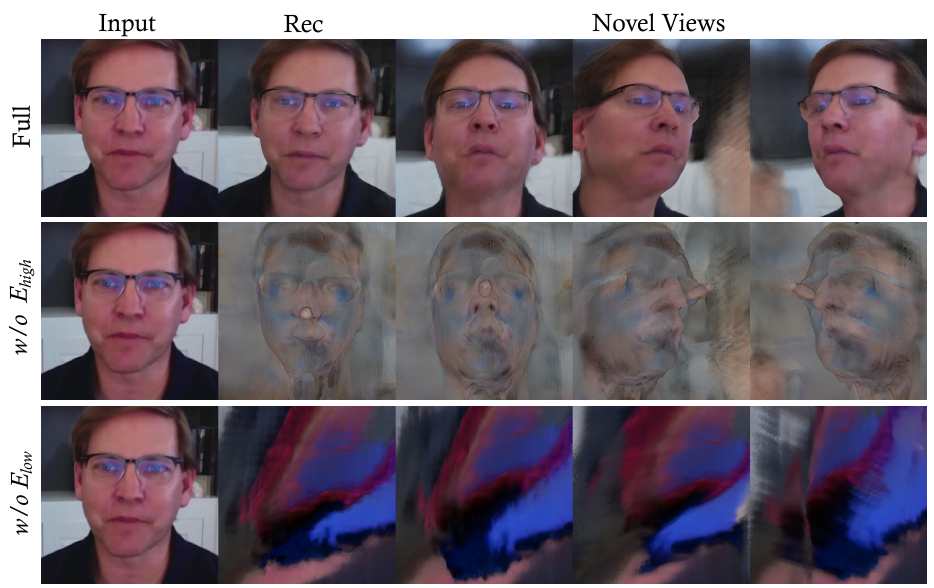}
  \caption{\textbf{Disentanglement in Live3D features.} We separately disable the features from $\mathrm{E}_{high}(\cdot)$ and $\mathrm{E}_{low}(\cdot)$ to infer the reconstructed image. Without $\mathrm{E}_{high}(\cdot)$, the output retains the coarse structure but loses its appearance. Conversely, when $\mathrm{E}_{low}(\cdot)$ is deactivated, the reconstructed portraits preserve the texture (such as the blue and purple reflection on the glasses) but fail to capture the geometry.}  
  \label{fig:toy}
\end{figure}

\subsection{Distilling Priors into a Lightweight Module}~\label{method:distill}

Our method aims to perform real-time 3D-aware editing for various prompts for the single-view portrait.
Thus, it needs powerful 3D knowledge for geometry reconstruction and editing prior to handling various control signals.
We leverage the strengths of both 3D GANs and diffusion models. 
In the following, we provide detailed presentations on how the knowledge of these two types of models is distilled into a lightweight module.

\noindent{\textbf{Feature Representation in Live3D}}. 
We carefully study the Live3D model and discover that,  as a two-branch triplane-based 3D reconstruction model, the low-resolution and high-resolution features from the Live3D encoder $\mathrm{E}_{low}(\cdot)$ and $\mathrm{E}_{high}(\cdot)$ tend to learn various levels of information without explicit guidance.
We conduct a study by disabling one of the branches and inferring the reconstructed images. 
As illustrated in Fig.~\ref{fig:toy}, when the high-resolution encoder $\mathrm{E}_{high}(\cdot)$ is disabled with random weights, the inference image retains a similar structure but loses its detailed appearance. 
Conversely, when the low-resolution encoder $\mathrm{E}_{low}(\cdot)$ is disabled, the reconstructed portraits preserve some of the texture from the input but struggle to capture the geometry.
Based on this analysis, the features of the two branches separately model low-frequency and high-frequency information. 
This insight motivates our model design in terms of how to distill knowledge from the diffusion model and 3D GANs.

\noindent{\textbf{Geometry Prediction for Inputs $\mathbf{I}$}}.
In our framework, we use the prompts $\mathbf{P}$ to refine the input portrait $\mathbf{I}$ into the edited image $\mathbf{I}_{p}$.
The $\mathbf{I}_{p}$ inherits a similar structure to the input $\mathbf{I}$, providing coarse geometry or structural guidance in the editing process.
As discussed in the above section, the encoder $\mathrm{E}_{low}(\cdot)$ consistently generates low-frequency features to represent geometry cues, making it well-suited for preserving structured information in the input portrait.
As a result, we freeze $\mathrm{E}_{low}(\cdot)$ and leverage it to produce structure features $\mathbf{F}_{g} = \mathrm{E}_{low}(\mathbf{I})$ for coarse geometry prediction.

\noindent{\textbf{Injecting Prompts $\mathbf{P}$ as Condition}}.
In contrast to input images, prompts typically offer more high-frequency and texture information to guide and control the editing process.
Accordingly, it is appropriate to incorporate the prompts into the high-level branch $\mathrm{E}_{high}(\cdot)$ in Live3D to extract the high-frequency features.
To incorporate prompts, we utilize the prompt encoder $\mathrm{E}_{p}(\cdot)$ to create prompt embeddings that are fused into our model through cross attention, which is similar to the strategy in Stable Diffusion~\cite{stablediffusion}.
Specifically, we add a cross-attention layer after $\mathrm{E}_{high}(\cdot)$ to obtain the feature updated with the prompt embeddings:
\begin{align}
    \mathbf{F}_{a} &= \mathrm{CrossAttention}(\mathrm{E}_{high}(\mathbf{I}), \mathrm{E}_{p}(\mathbf{P})),
\end{align}
where the encoder $\mathbf{E}_{p}(\cdot)$ is a transformer (MAE~\cite{he2022masked} for image and CLIP~\cite{clip} for text).
The $\mathrm{E}_{high}(\mathbf{I})$ serves as the query and $\mathrm{E}_{p}(\mathbf{P})$ is employed as key and value.
The updated feature $\mathbf{F}_{a}$ and the geometry feature $\mathbf{F}_{g}$ are then feed into the decoder $\mathrm{E}_{t}(\cdot)$ to infer the triplane $\mathbf{T}_{\mathbf{p}}$:
\begin{align}
    \mathbf{T}_{\mathbf{p}} = \mathrm{E}_{t}(\mathbf{F}_{a}, \mathbf{F}_{g}).
\end{align}

\noindent{\textbf{Distilling Diffusion and 3D GAN Prior}}.
Unlike previous methods that require large amounts of 3D data, our model operates with only 2D paired data generated by the 2D editing model~\cite{instructpix2pix}.
For both visual and textual prompts, we generate edited images $\mathbf{I}_{gt}$ with the diffusion-based image editing model~\cite{instructpix2pix} as the pseudo labels. 
The details of data generation can be found in~\cref{subsec:experimental_setup}.

With the predicted triplane $\mathbf{T}_{\mathbf{p}}$, we can render the image using the pretrained EG3D model and then calculate the reconstruction loss as follows:
\begin{align}
    \mathcal{L}_{2d} = \ell_\mathbf{I}(\mathbf{I}_{gt}, \mathrm{R}(\mathbf{T_{\mathbf{p}}}, \mathbf{c})),
\end{align}
where $\mathrm{R}(\cdot)$ is the rendering module of EG3D, $\mathbf{c}$ is the camera pose of input portrait, and $\ell_1(\cdot)$ is an image reconstruction loss penalizing the difference between the ground truth $\mathbf{I}_{gt}$ and the rendering $\mathbf{I}_{p} = \mathrm{R}(\mathbf{T_{\mathbf{p}}}, \mathbf{c}))$.
It's important to note that $\mathcal{L}_{2d}$, used to reconstruct the $\mathbf{I}_{gt}$, essentially distills the knowledge of editing from the diffusion model.

We observe that $\mathcal{L}_{2d}$ helps the model to reconstruct well on the input camera view but suffers from geometry distortion during novel-view synthesis.
To fully exploit the 3D properties of Live3D, we propose to distill the 3D knowledge of Live3D. 
Specifically, we leverage the pretrained Live3D to infer the triplane $\mathbf{T}_{gt}$, and multi-view depths $\mathcal{D}_{gt}$ and images $\mathcal{I}_{gt}$ of the pseudo-label image $\mathbf{I}_{gt}$:
\begin{align}
\mathbf{T}_{gt}, \mathcal{D}_{gt} = \mathrm{G}(\mathbf{I}_{gt}, \mathcal{C}),
\end{align}
where $\mathrm{G}(\cdot)$ represents the inference process of Live3D, where $\mathcal{C} = \{\mathbf{c}_1, .. , \mathbf{c}_n\}$ is the camera set, and $n$ denotes the number of cameras.
We also render the multi-view images $\mathcal{I}_{p}$ and depths $\mathcal{D}_{p}$ of the edited image $\mathrm{I}_{p}$ from the triplane $\mathbf{T}_{p}$ and then define the objective:
\begin{align}
\mathcal{L}_{3d} = \ell_{\mathcal{I}}(\mathcal{I}_{p}, \mathcal{I}_{gt}) + \ell_{\mathbf{T}}(\mathbf{T}_{p}, \mathbf{T}_{gt}) + \ell_{\mathcal{D}}(\mathcal{D}_{p}, \mathcal{D}_{gt}),
\end{align}
where $\ell_{\mathcal{I}}(\cdot)$, $\ell_{\mathbf{T}}(\cdot)$ and $\ell_{\mathcal{D}}(\cdot)$ is the reconstruction loss penalizing the difference in image, triplane, and depth between $\mathrm{I}_{p}$ and $\mathrm{I}_{gt}$.

\subsection{Training and Inference}~\label{method:training}
\label{subsec:training_and_inference}

\noindent{\textbf{Training}}.
During the training phase, we sample a triplet consisting of the input portrait $\mathbf{I}$ along with its camera pose $\mathbf{c}$, prompts $\mathbf{P}$, and the pseudo-label image $\mathbf{I}_{gt}$ generated by the 2D diffusion model.
We leverage the Live3D model as a pretrained model and add an additional prompts encoder $\mathrm{E}_{p}$ to extract prompt embeddings. The overall learning objective can be described as follows:
\begin{align}
L = \lambda_{1}L_{2D} + \lambda_{2}L_{3D}, \label{eq:loss}
\end{align}
where $\lambda_{1}$ and $\lambda_{2}$ are loss weights. In our setting, $\lambda_{1}$ and $\lambda_{2}$ are both set to 1.0.
For the reconstruction loss, $\ell_{\mathcal{I}}(\cdot)$, $\ell_{\mathrm{I}}(\cdot)$ are combinations of L$_{2}$ loss and LPIPS loss~\cite{lpips}, with loss weights being 1 and 2, respectively. 
$\ell_{\mathcal{D}}(\cdot)$ and $\ell_{\mathrm{T}}(\cdot)$ are L$_{1}$ loss.
Notably, based on the study on Live3D feature representation, during training only $\mathrm{E}_{p}(\cdot)$, $\mathrm{E}_t(\cdot)$ are set to be learnable, while other modules are frozen. 
It allows our model to leverage Live3D knowledge as much as possible and converge at a very fast speed.
 
\noindent{\textbf{Inference}}.
For inference, users can provide a single 2D portrait image and choose an image prompt or text prompt.
Our model is able to generate the edited 3D NeRF along with photorealistic view synthesis.

\noindent{\textbf{Customized Prompts Adaptation}}.
To accommodate customized styles provided by users, we propose a method to adapt our pretrained encoder to user-defined styles. 
We increase the tuning efficiency by optimizing only $\mathrm{E}_{p}(\cdot)$ and the normalization layers in $\mathrm{E}_{t}(\cdot)$ with the same learning objective Eq.~\ref{eq:loss}.
This method allows us to limit the training data to only 10 image pairs and the learning time to 5 minutes on a single GPU.

%% file: sections/4.experiments.tex
\section{Experiments}
\label{sec:exp}

\subsection{Experimental Setup}
\label{subsec:experimental_setup}

\noindent{\textbf{Training Settings}}.
For real faces, we adopt the FFHQ dataset~\cite{karras2019style} at $512\times512$ resolution with camera parameters aligned by EG3D~\cite{eg3d}.
In order to obtain the stylized images as pseudo labels, and the corresponding textual and visual prompts, we leverage the text-guided editing model InstructPix2Pix (IP2P)~\cite{instructpix2pix} to edit the real faces. 
Specifically, we manually generate and select textual prompts, and send them to IP2P together with the real images to obtain the edited 2D images.
The text prompts of our training and testing set are obtained by augmenting the aforementioned textual prompts with the large language model GPT~\cite{brown2020language} following IP2P~\cite{instructpix2pix}.
While the image prompts of our training and testing set are picked from the edited 2D images.
We conduct experiments on 20 styles, and for each style, we synthesize 1000 images, resulting in 20,000 image pairs with text prompts for model training.
For each style, we use 8 textual prompts for training and 5 textual prompts for testing.
We adopt a learning rate of 5e-5 and optimize the model for  60k iterations with a batch size of 32.
The entire training procedure is completed over a period of 40 hours using 8 NVIDIA A100 GPUs.

\begin{figure}[t]
    \centering
    \includegraphics[width=1.0\textwidth]{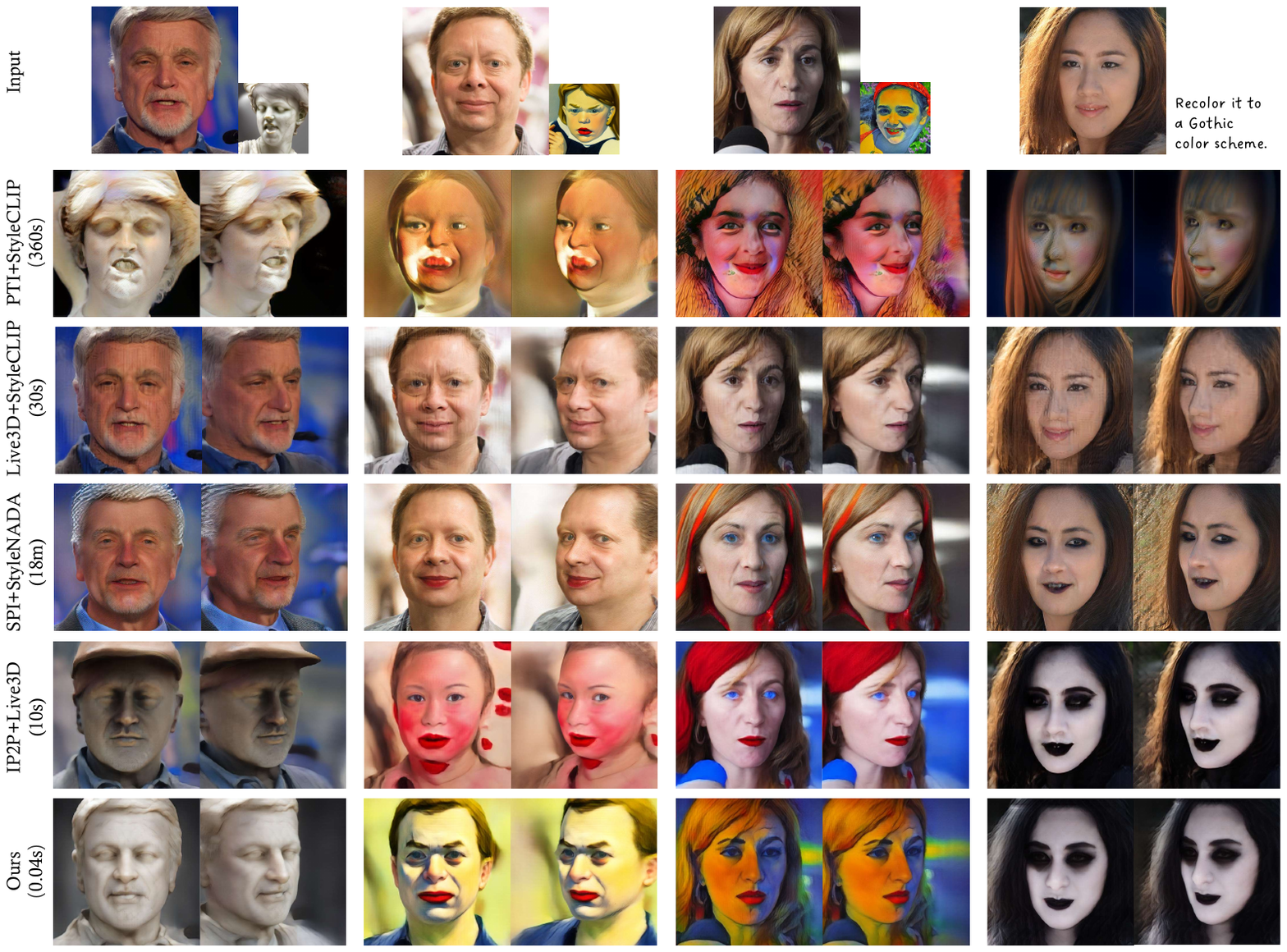}
    \caption{\textbf{Qualitative comparisons.}  We compare the results of several baselines with image prompts and text prompts. In each case, we include the edited portraits as well as their novel view renderings. Our method generates high-quality edited portraits with better 3D quality and alignment with the referenced prompts.}
    \label{fig:qualitative}
\end{figure}

\noindent{\textbf{Baselines}}.
We conduct comparisons against two categories of methods that achieve analogous outcomes: 1) 3D reconstruction coupled with 3D editing; 2) 2D editing followed by 3D reconstruction.
For the first category, we implement three baseline approaches, wherein we firstly leverage PTI~\cite{pti} or Live3D~\cite{live3d} for reconstruction and edit both of them following StyleCLIP~\cite{styleclip}.
Specifically, for experiments in PTI + StyleCLIP, we first conduct pivotal tuning~\cite{pti} based on the predicted latent codes predicted by e4e~\cite{tov2021e4e} to reconstruct the given image in a 3D-aware manner.
Then we perform editing on the latent codes following~\cite{styleclip}, where CLIP loss~\cite{clip} is utilized between the model output and the prompt (no matter whether it is an image or text).
The setting of Live3D + StyleCLIP is similar to PTI + StyleCLIP except that we obtain the triplane by Live3D and perform editing on this 3D representation.
SPI~\cite{yin20233d} + StyleGAN-NADA~\cite{gal2022stylegannada} (StyleNADA for the abbreviation in~\cref{fig:qualitative}) serves as the third baseline in the first category, where SPI is used for 3D reconstruction and StyleGAN-NADA is utilized to optimize the 3D generator for editing.
In the second category, we employ a diffusion-based text-guided 2D image editing method IP2P~\cite{instructpix2pix} prior to executing 3D reconstruction via Live3D. 
IP2P being a text-instructed editing model, we need to first extract text prompts from the reference image with the image caption algorithm BLIP~\cite{li2022blip} when image prompts are input.
The 2D edited results could then be obtained by sending the input image and the text prompt to IP2P.
The final 3D results are acquired by inverting the 2D editing results with Live3D.

\noindent{\textbf{Evaluation Criteria}}.
We evaluate all metrics on 100 pairs of images processed from FFHQ.
We conduct a comprehensive evaluation of the editing performance in the following four aspects: identity preservation, reference alignment, 3D consistency, and inference speed. 
\textit{\textbf{1) Identity preservation ($\mathbf{ID_t}$)}} aims to measure the preservation of the original identity by calculating the cosine similarity between the identity feature of the input image $\mathbf{I}$ and that of the edited image $\mathbf{I}_{p}$. We use the ArcFace model~\cite{arcface} to extract identity features. 
\textit{\textbf{2) Reference alignment ($\mathbf{CLIP_r}$)}} targets at assessing the alignment of the output editing styles to the desired input prompt by computing the cosine similarity in the CLIP~\cite{styleclip} feature space.
\textit{\textbf{3) 3D consistency (3D)}} on the edited outputs is measured following the evaluation protocols established by EG3D~\cite{eg3d}. This involves calculating the identity similarity across multiple views.
\textit{\textbf{4) Inference speed (Time)}} is measured on a single NVIDIA A6000 GPU with an average of 100 samples.

\input{tables/quantitative}

\begin{figure}[t]
    \centering
    \includegraphics[width=1.0\textwidth]{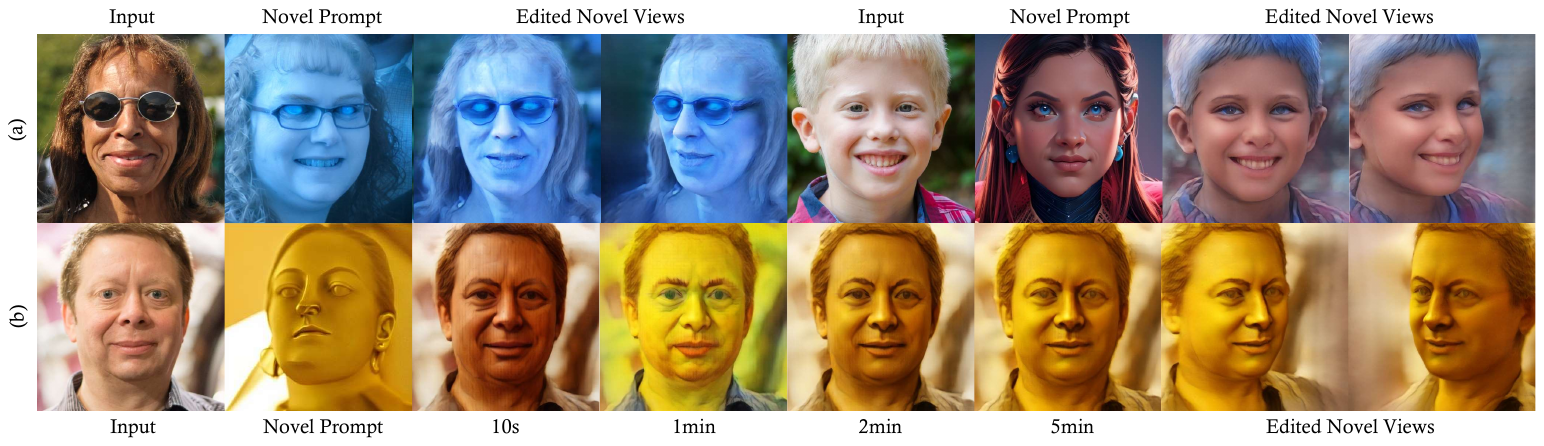}
    \caption{
    \textbf{(a) Generated results of customized prompt adaptation and (b) its learning process}. 
    We show the intermediate testing results at 10s, 1min, 2min and 5min during adaptation for the novel style golden statue.}
    \label{fig:novel_style_adaptation}
\end{figure}

\begin{figure}[t]
\centering
\includegraphics[width=1.0\linewidth]{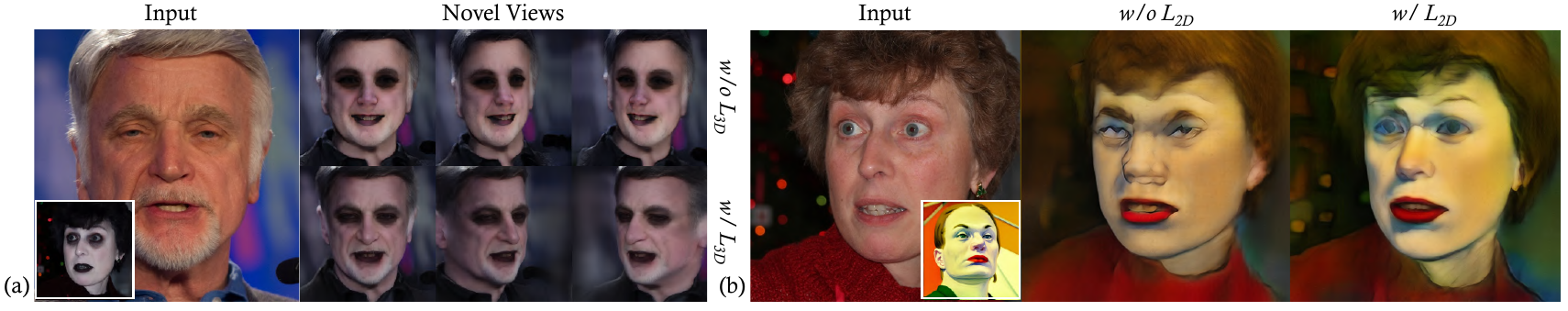}
  \caption{\textbf{Qualitative comparison for ablations on} (a) the distillation loss of 3D GAN ($\mathcal{L}_{3D}$) and (b) diffusion models ($\mathcal{L}_{2D}$).}
  \label{fig:ablation}
\end{figure}

\subsection{Efficient 3D-aware Portrait Editing}
We make an in-depth analysis of our efficient portrait editing method both quantitatively and qualitatively.
The results are included in Tab.~\ref{tab:quantitative} and Fig.~\ref{fig:qualitative}.
For reference alignment, we do not report $CLIP_{r}$ for methods that leverage CLIP for optimization since it is evaluated with CLIP as well.
The standout feature of our method is its efficiency.
Our method achieves an inference speed of merely 40ms, which improves over 100 times compared to the fastest existing baselines, which require around 10 seconds.
Because of the efficient knowledge distillation, our approach is also good in preserving the identity, adhering more closely to the reference prompts and achieving the best 3D consistency.

As in Fig.~\ref{fig:qualitative}, compared with PTI + CLIP methods, our model enables precise alignment with inputs and prompts, while the baseline always generates low-fidelity textures, and the resulting geometry has many artifacts.
Although Live3D + CLIP achieves the best $ID_{t}$ score, the edited results are basically unchanged compared to the input.
We suspect that Live3D does not possess a latent space and loses the editing priors, making it challenging for the results of CLIP optimization to align well with prompts.
SPI + StyleGAN-NADA performs relatively slowly (18 minutes per instance) because both the 3D reconstruction and the editing are achieved by optimizing the generator. 
Although it can produce realistic textures, it fails to reconstruct fine details and to edit following the prompts.
In contrast to the 2D editing and subsequent 3D reconstruction pipeline (IP2P + Live3D), our method produces textures more consistent with the prompt and aligns better with the input structure.

\subsection{Adaptation for Customized Editing} 
With the trained editing network, our method already supports a variety of 3D-aware editing styles. 
Following DreamBooth~\cite{dreambooth} and to expand the selection and better conform to users' preferences, we offer an efficient method for fast adaptation to the user-defined customized prompts.
Users can personalize the editing network by providing a modest set of 10 reference editing pairs.
These pairs can be either handpicked from artist-created examples or produced using text-guided image editing models. 
As described in~\cref{subsec:training_and_inference}, only the partial of the encoder would be optimized.
The adaptation process itself is remarkably fast, requiring only about 5 minutes to accomplish the learning of the customized prompts.
Please also note that, once the user-defined style has been trained, all the incoming testing results can be obtained in a real-time manner.
In Fig.~\ref{fig:novel_style_adaptation}, we present the testing results and learning process of the adaptation.
Our model can quickly master the customized knowledge in about 2 minutes.
With further training (\textit{e.g.}, 5 minutes), the edited results become more stylized, demonstrating a trade-off between authenticity and stylization.
Upon completion of the adaptation, the method allows users to edit testing inputs in these newly learned styles with a minimal inference time of 0.04s. 
This rapid performance indicates that the method is well-optimized for real-time applications, providing a seamless and efficient user experience.

\input{tables/ablation_pretrain}
\input{tables/ablation_adaptation}

\begin{figure}[t]
\centering
\includegraphics[width=0.95\linewidth]{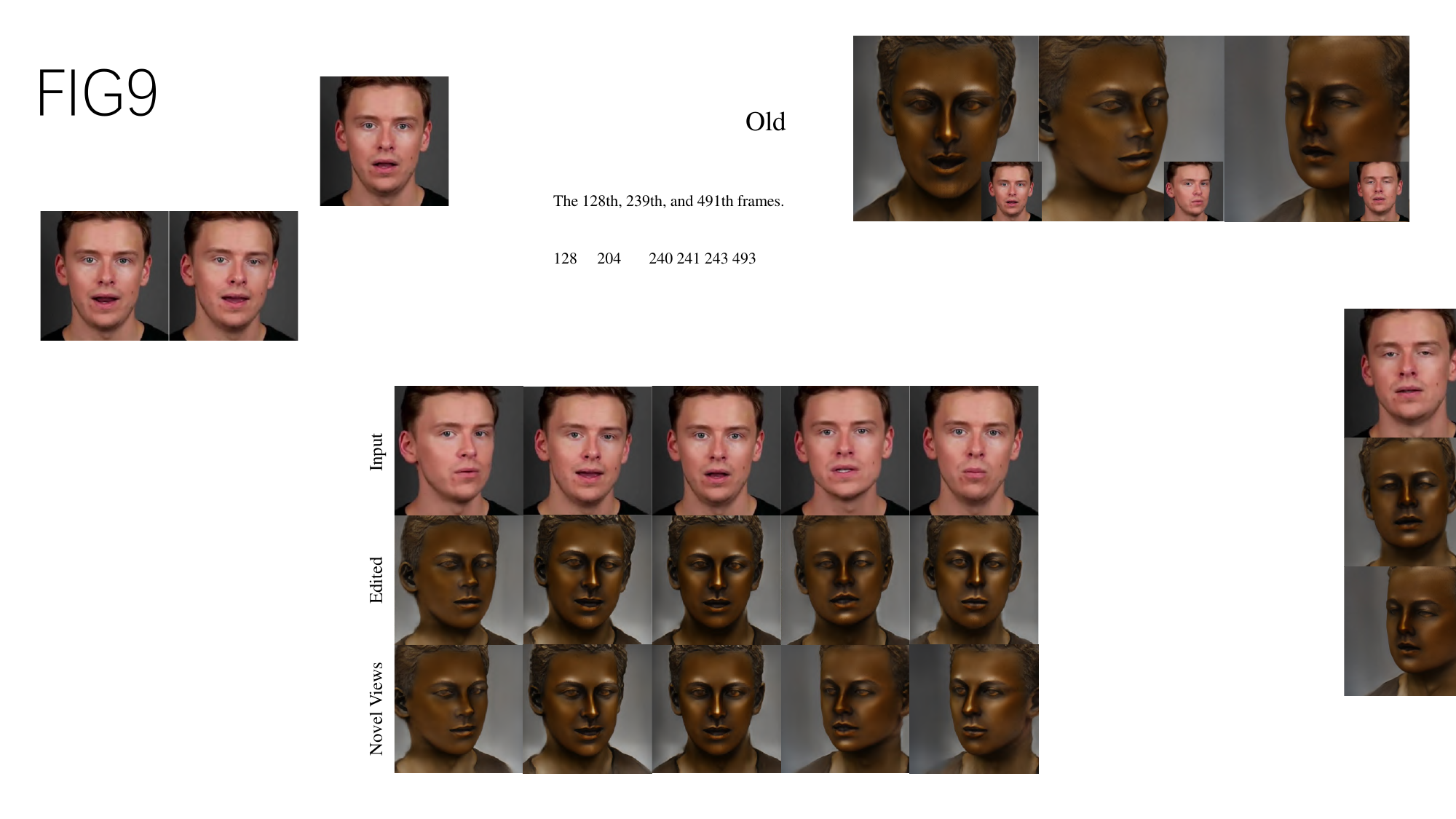}
  \caption{ \textbf{Video editing results with \method}. We use text prompts "Make him a bronze statue" to edit the input video. Our method can accurately reconstruct and preserve challenging facial expressions and achieve high-quality novel view renderings.}
  \label{fig:video}
\end{figure}

\subsection{Ablation Studies}
We analyze our model and validate our design choice by ablating the components.

\noindent{\textbf{$\mathcal{L}_{3D}$ for Distillation of Live3D}}. 
We compare our model trained with and without $\mathcal{L}_{3D}$. 
As shown in Fig.~\ref{fig:ablation}, we observe that the model without $\mathcal{L}_{3D}$ results in very flat geometry, and the synthesis of the side view loses the normal structure with many artifacts. 
This highlights the importance of the 3D knowledge from Live3D in training our model with 2D paired data, enabling the inference of reasonable face geometry without relying on any 3D data sources.

\noindent{\textbf{$\mathcal{L}_{2D}$ for Distillation of Diffusion Models}}.
We also study the effect of $\mathcal{L}_{2D}$, which represents using 2D editing priors from the diffusion model~\cite{instructpix2pix} to optimize the encoder.
From Fig.~\ref{fig:ablation}, we observe that $\mathcal{L}_{2D}$ is critical for our model, especially in preserving detailed texture. 
Without it, the edited image tends to lose the appearance information from prompts and exhibits structural artifacts around the nose.

\noindent\textbf{{Ablation on Pretraining Data Scale for Customized Prompt Adaptations}}. 
In this part, we conduct ablation studies on the pretraining scale and investigate its influence on customized propmts.
Specifically, we choose various 3DPE models pretrained with different style amounts (3, 10, and 20 styles respectively) and perform novel prompt adaptation on a single A6000 GPU.
We choose the golden statue as the customized style and report LPIPS score between the predicted stylized images and the pseudo stylized labels produced by IP2P~\cite{instructpix2pix}.
Experimental results in~\cref{tab:abaltion_pretrain_data_scale} reveal that the proposed method could benefit from the improving pretraining data scale and adapt faster and eventually perform better with more data, which suggests the potential of the model for future scaling up.

\noindent\textbf{{Ablation on Tuning Data Scale for Customized Prompt Adaptations}}. 
To investigate the ability of our method to adapt to user-defined novel prompts, we conduct an ablation study on the number of novel-prompt data pairs required for training.
Similar to the prior experiments on the pretraining scale, we perform adaptation with 2, 5, 10, 20, or 50 pairs of novel-prompt data on a single A6000 GPU.
We experimentally find that our model can converge in about 5 minutes, and therefore, we report the perceptual distance between the model output and the provided ground truth at 0.1min, 1min, 2min, and 5min for comparison.
As shown in Tab. \ref{tab:abaltion_adaptation_data_scale}, our model can adapt to the new style in around 2 minutes. 
As the number of provided data pairs increases, the model can adapt faster at the beginning.

\subsection{Applications} 

One application of our pipeline is to edit videos, requiring accurate reconstruction and efficient editing simultaneously. 
We demonstrate the effectiveness of our pipeline on talking face videos, as illustrated in Fig.~\ref{fig:video}. 
In this experiment, we perform sequence editing in a frame-by-frame manner by sending an image sequence from TalkingHead-1KH~\cite{wang2021talkinghead1kh} dataset together with the text prompt "Make him a bronze statue" to \method.
The three rows in Fig.~\ref{fig:video} respectively illustrate the input frames, the edited frames, and the novel views.
Our method accurately reconstructs facial expressions, while achieving high-quality editing results. 
Since our method is 3D-aware, we can naturally view videos from novel perspectives. 
To better experience our method, we design an interactive system that allows users for interactive editing.
Users are required to provide the image or text prompt to indicate the desired style and input an input image that is to be edited, and the system will apply our method on the given image and prompt to output the edited image as well as novel-view edited results.

%% file: tables/quantitative.tex
\begin{table}[t]
  \caption{\textbf{Quantitative comparisons}. We compare several baselines on the 100 images of FFHQ dataset. It's important to note that we exclude $CLIP_{r}$ for PTI+StyleCLIP and Live3D+StyleCLIP since these models utilize CLIP for optimization. Our model excels in 3D quality and achieves a remarkable 250x improvement (compared with IP2P+Live3D) in inference speed, achieving real-time performance.
  }
  \label{tab:quantitative}
  \scriptsize
  \setlength{\tabcolsep}{12pt}
  \centering
  \begin{tabular}{lccccc}
    \toprule
    Method  & $ID_{t}\uparrow$  & $CLIP_{r}\uparrow$ & 3D$\uparrow$ & Time$\downarrow$\\
    \midrule
    PTI+StyleCLIP & 0.11  & -  & 0.73 & 360s \\
    Live3D+StyleCLIP & \textbf{0.63}  & -  & 0.72 & 30s \\
    \midrule
    IP2P+Live3D & 0.44  & 0.62  & 0.75 & 10s \\
    \midrule 
    Ours & 0.47 & \textbf{0.73}  & \textbf{0.76} & \textbf{0.04s} \\
    \bottomrule
  \end{tabular}
\end{table}

%% file: tables/ablation_pretrain.tex
\begin{table}[t]
  \caption{
    \textbf{Ablation study on the number of styles used for pretraining}. We report LPIPS score after 0.1min, 1min, 2min, and 5min fine-tuning for evaluation.
  }
  \label{tab:abaltion_pretrain_data_scale}
  \setlength{\tabcolsep}{12pt}
  \centering
  \scriptsize
  \begin{tabular}{cccccc}
    \toprule
    \#Style/Time  & 0.1min & 1min & 2min & 5min\\
    \midrule
    3 & 0.7758 &  0.6370  & 0.5312  & 0.5220\\
    10 &  0.5932 &  0.5734  & 0.5106 & 0.5012 \\
    20 & 0.5703 & 0.5290 &   0.5022 & 0.4849\\
    \bottomrule
  \end{tabular}
\end{table}

%% file: tables/ablation_adaptation.tex
\begin{table}[t]
  \caption{
    \textbf{Ablation study on the number of data pairs used for novel prompt adaptation}. We report LPIPS score after 0.1min, 1min, 2min, and 5min fine-tuning for evaluation.
  }
  \label{tab:abaltion_adaptation_data_scale}
  \setlength{\tabcolsep}{12pt}
  \centering
  \scriptsize
  \begin{tabular}{cccccc}
    \toprule
    \#Pair/Time  & 0.1min & 1min & 2min & 5min\\
    \midrule
    2 & 0.6191 &  0.5431  & 0.5213  & 0.5212\\
    5 &  0.6148 &  0.5324  & 0.5064 & 0.4971 \\
    10 &  0.6076 &  0.5242  & 0.4983 & 0.4935 \\
    20 & 0.5981 & 0.5525 &   0.4954 & 0.4895\\
    50 & 0.5891 &  0.5718 & 0.4922  & 0.4797\\
    \bottomrule
  \end{tabular}
\end{table}

%% file: sections/5.conclusion.tex
\section{Discussion}

\noindent{\textbf{Limitations.}}
Despite achieving state-of-the-art performance in quality and efficiency, our method exhibits inconsistencies in details for novel-view rendering because the EG3D framework relies on a super-resolution module.
Additionally, when our method is applied to video editing, it presents flickering artifacts since our model is originally designed for image-based editing.

\noindent{\textbf{Conclusion.}}
In this work, we introduce \method for real-time 3D-aware portrait editing. 
By distilling the powerful knowledge of diffusion models and 3D GANs into a lightweight module, our model significantly reduces editing time while ensuring quality. 
Benefiting from the distillation process, \method could handle multiple styles in a single model and perform both image-guided and text-guided editing.
The advantages of our method empower us to generate photorealistic 3D portraits, a capability crucial for the visual effects industry, AR/VR systems, and teleconferencing, among other applications, while we firmly oppose any misuse that could compromise privacy and security.